# Modeling biological face recognition with deep convolutional neural networks

**Leonard Elia van Dyck[1,2]\* and Walter Roland Gruber[1,2]**

[1]Department of Psychology, University of Salzburg, Austria
[2]Centre for Cognitive Neuroscience, University of Salzburg, Austria

**\*Correspondence:**
Leonard Elia van Dyck
lenny.vandyck@gmail.com

## Abstract

Deep convolutional neural networks (DCNNs) have become the state-of-the-art computational models of biological object recognition. Their remarkable success has helped vision science break new ground and recent efforts have started to transfer this achievement to research on biological face recognition. In this regard, face detection can be investigated by comparing face-selective biological neurons and brain areas to artificial neurons and model layers. Similarly, face identification can be examined by comparing in vivo and in silico multidimensional "face spaces". In this review, we summarize the first studies that use DCNNs to model biological face recognition. On the basis of a broad spectrum of behavioral and computational evidence, we conclude that DCNNs are useful models that closely resemble the general hierarchical organization of face recognition in the ventral visual pathway and the core face network. In two exemplary spotlights, we emphasize the unique scientific contributions of these models. First, studies on face detection in DCNNs indicate that elementary face selectivity emerges automatically through feedforward processing even in the absence of visual experience. Second, studies on face identification in DCNNs suggest that identity-specific experience and generative mechanisms facilitate this particular challenge. Taken together, as this novel modeling approach enables close control of predisposition (i.e., architecture) and experience (i.e., training data), it may be suited to inform long-standing debates on the substrates of biological face recognition.

# Introduction

## Face recognition in the brain

Since decades, neuroscientists and psychologists are intrigued by the behavioral and computational processes underlying biological face recognition. As especially faces of humans and nonhuman primates carry important social information (Freiwald et al., 2016), their perception could have developed as a highly specialized and effective form of object recognition (Kanwisher, 2000). Faces are processed in terms of physical and functional properties (Bruce & Young, 1986). On one side, face detection enables spotting inter-class differences between faces and other objects and is often studied in the context of neuronal face selectivity (Tsao & Livingstone, 2008), that is, the functional specialization or "preference" of neurons for faces compared to a wide spectrum of other objects. On the other side, face identification relates to intra-class differences between individual face examples and involves numerous dimensional features that span a representational "face space" encoding the similarity of different identities, that is, their position in this multidimensional space in relation to examples of other identities (Valentine et al., 2016). In this review, we illustrate how latest studies foreshadow that state-of-the-art computational encoding models, namely deep convolutional neural networks (hereafter DCNNs), could be used to decipher biological face recognition.

A remarkable collection of neuroscientific studies has long identified face-selective neurons (Desimone, 1991; Perrett et al., 1992; Tsao et al., 2006) and brain areas (Kanwisher et al., 1997; Kanwisher & Yovel, 2006; Perrett et al., 1987) as the neural basis for an assumed specialized treatment of face configurations (Freiwald, 2020; Freiwald & Tsao, 2010; Hesse & Tsao, 2020). Studies using functional magnetic resonance imaging (fMRI) have discovered that biological face recognition involves distributed networks of face-selective areas. This core face network comprises the

**Modeling face recognition with DCNNs**

fusiform face area, occipital face area, and posterior superior temporal sulcus (Haxby et al., 2000; Kanwisher et al., 1997), and is supported by additional regions. The question of whether face selectivity arises through predisposition from innate mechanisms in the form of an automatic face bias (Johnson et al., 1991; Johnson & Mareschal, 2001), is acquired through experience leading to face expertise (Gauthier & Bukach, 2007; Gauthier & Nelson, 2001; Young & Burton, 2018), or an interaction of both (Srihasam et al., 2012; Srihasam et al., 2014), has sparked heated debates. Recent findings have steered this discussion in a completely new direction by indicating that faces may not be "special" after all, as their selectivity could arise as a byproduct of domain-general features (Bao et al., 2020; Vinken et al., 2022) or optimization for a domain-specific task (Dobs et al., 2022b; Kanwisher et al., 2023a; Yovel et al., 2022b).

From a theoretical standpoint, the endeavor to disentangle the influences of nature and nurture on face recognition ultimately calls for more impactful and controlled experimental manipulations such as lesions, deprivations, or extensive exposure. However, from a practical standpoint, these designs are challenging and unethical to study in living beings and would not allow the examination of the brain's entire face processing system at once. Consequently, recent models from the field of deep learning could provide a suitable alternative by simulating such experiments in silico.

**DCNNs as models of biological object recognition**

DCNNs (e.g., Krizhevsky et al., 2012; LeCun et al., 2015) are a subset of artificial neural networks and commonly used in the field of computer vision to solve a wide range of image classification tasks. The seminal discovery of simple and complex cells in the visual cortex (Hubel & Wiesel, 1959, 1962) once inspired these architectures, which stack convolutional and pooling operations to perform a series of linear and nonlinear transformations of input data through multiple layers of learnable filters. Since their invention, DCNNs have been rapidly adopted in the field of visual neuroscience. Today, because of their strong architectural and functional similarities to the ventral visual pathway, particularly in humans and nonhuman primates, DCNNs are established as state-of-the-art computational models of biological object recognition (Cadieu et al., 2014; Cichy et al., 2016; Greene & Hansen, 2018; Güçlü & van Gerven, 2015; Khaligh-Razavi & Kriegeskorte, 2014; Kheradpisheh et al., 2016; Yamins et al., 2013; Yamins et al., 2014).

However, the general enthusiasm for these promising computational models has also raised critical voices from opponents of this research agenda. A part of this criticism relates directly to DCNNs and primarily addresses the shortcomings of the models, such as their need for large amounts of labeled data, their lack of generalization, their assumption of a relatively stable world, their lack of transparency as a "black box", and their reliance on fundamentally different processing mechanisms (e.g., Leek et al., 2022; Marcus, 2018). Another part of this criticism relates to the current research program that praises DCNNs as the best models of human object recognition. In a recent critical article, Bowers et al. (2022) point out various principled and practical problems of this research program and argue that it relies heavily on "prediction-based experiments", which focus on blindly explaining the highest possible variance rather than examining the actual underlying effects. They reason that, in contrast to "controlled experiments", this approach introduces various questionable practices in the field, such as biases in the selection of architectures, datasets, and parameters, as well as neglect of alternative explanations and hypotheses. In essence, they contend that DCNNs "*share little overlap with biological vision*" and "*account for almost no results from psychological research*" (Bowers et al., 2022).

Conversely, proponents of this research agenda argue that DCNNs are sophisticated models of biological vision that can be used to reveal novel and otherwise unobtainable insights. In a recent definition of this endeavor, Doerig et al. (2022) advocate that studies using artificial neural networks are often much like "controlled experiments", because they allow for close control over both the effects of architecture and experience, while even avoiding some of the limitations and biases inherent in classical paradigms. Following this chain of reasoning, DCNNs can be used to effectively investigate multiple levels, including behavior, computation, single units to complex dynamics, natural sensory information, naturalistic environments, and developmental processes. According to this view, dissimilarities between brains and artificial neural networks should not immediately refute the use of the latter, but rather point to intriguing mechanisms and causalities that require further exploration.

## Modeling face recognition with DCNNs

Therefore, artificial neural networks provide a balance between the computational abstraction required and the biological detail needed to test complex hypotheses about the brain. Importantly, DCNNs can also be used to evaluate causal hypotheses that go far beyond "prediction-based experiments" by controlling factors that influence optimization for a particular task, thus isolating the minimum sufficient requirements for a specific phenomenon to occur (Kanwisher et al., 2023b). As their architecture, initial training, subsequent fine-tuning, and artificial neurons can be closely controlled and directly manipulated, DCNNs may provide a useful addition to the field of face perception research as encoding models of face recognition (reviewed here; see Figure 1) and decoding models for face reconstruction from neural data (VanRullen & Reddy, 2019). On the basis of this account, it seems highly reasonable to replace the brain as a targeted "black box" with a more accessible and controllable DCNN as a modeled "transparent box" that is still sufficiently complex but certainly easier to study.

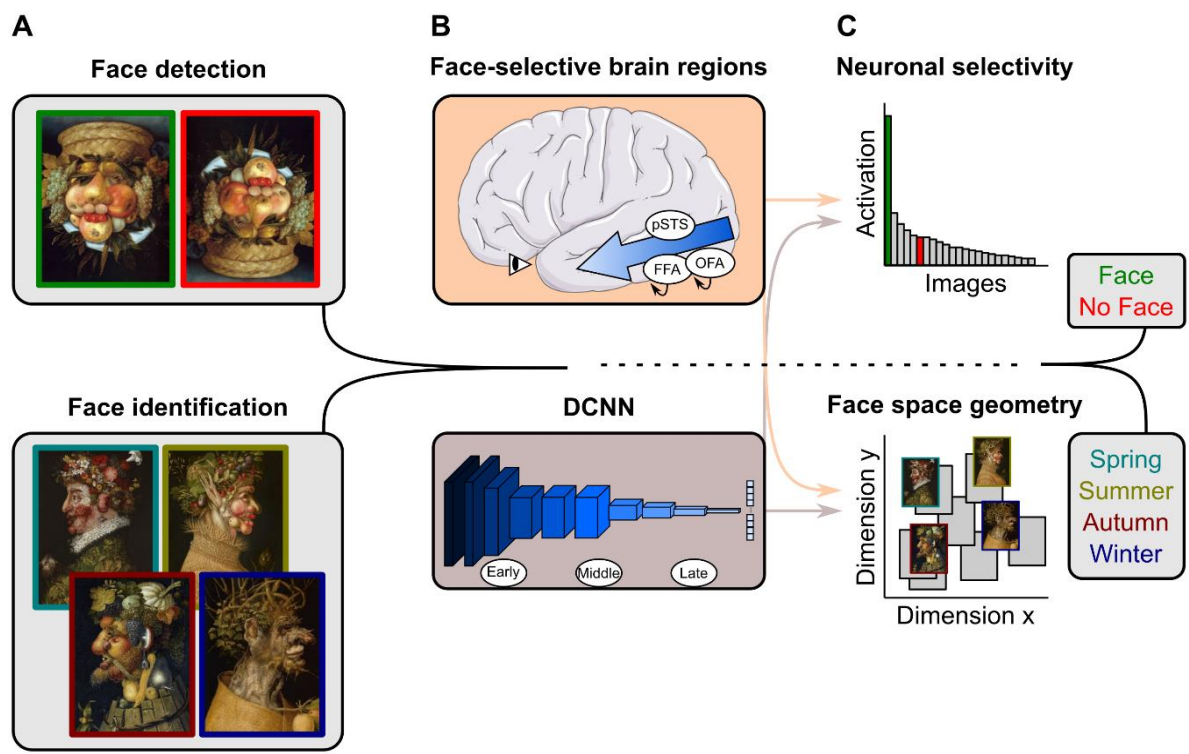

**Figure 1. Face recognition in the brain and DCNNs. (A)** Face detection and identification require recognition of inter-class differences (i.e., faces and other objects) and intra-class differences (i.e., faces of various identities), respectively. **(B)** The brain contains many face-selective biological neurons and regions such as fusiform face area, occipital face area, and posterior superior temporal sulcus. DCNNs are able to develop similar face-selective artificial neurons throughout their early, middle, and late

**Modeling face recognition with DCNNs**

layers. **(C)** Face detection is often studied in the context of neuronal selectivity, which quantifies how much a given neuron is activated by faces compared with other objects. Face identification is usually investigated in the context of a multidimensional "face space", which encodes the similarity of identities along various dimensional features. Both investigations can be performed for biological and artificial neurons alike. **Source:** Paintings by Giuseppe Arcimboldo are for demonstration purposes only and publicly available at https://commons.wikimedia.org. Parts of the figure were adapted from SMART Servier Medical Art, licensed under a Creative Commons Attribution 3.0 unported license.

This perspective provides a suitable starting point to explore the literature on modeling biological face recognition with DCNNs. With this in mind, our review is organized as follows: To begin with, we will summarize the behavioral and computational evidence supporting the idea that DCNNs are also useful models for biological face recognition, which involves both detection and identification. Then, we will outline two topical examples of how this modeling approach is being used to gain new insights. As we will see in a first spotlight, recent studies indicate that elementary face selectivity may arise automatically through feedforward hierarchies and even without visual experience. As we will see in a second spotlight, recent evidence also suggests that face identification is enhanced by identity-specific experience and additional generative mechanisms. Finally, we will revisit the results of the initial studies and discuss open questions and future directions of this rapidly evolving field.

## DCNNs as models of biological face recognition

*Are DCNNs actually useful models to study biological face recognition?*

To answer this question, we will evaluate the current evidence to determine whether DCNNs provide an adequate explanation for the different stages of this ability and thus constitute a suitable neuroscientific model (Doerig et al., 2022). We will focus primarily on the behavioral and computational levels (including findings on single units and complex dynamics), as there exists a considerable amount of literature in these areas. In addition, it should be noted that most studies aim to examine natural sensory information (e.g., neural data) recorded in naturalistic environments (e.g., naturalistic face images). For consistency in terminology, we will refer to DCNNs trained only on face images for face identification as "face-id models", trained on face and object images for face detection as "face-de models", trained only on nonface object images for object categorization as "object-cat models", and untrained randomly initialized as "untrained models".

## Behavioral level

At the behavioral level, comparisons between humans and DCNNs usually involve the evaluation of categorization accuracies, error patterns, similarity ratings, and their underlying representations in response to face stimuli under naturalistic and/or manipulated conditions.

*Do humans and DCNNs perform similarly in face recognition?*

DCNNs have long matched human performance in face detection and identification (e.g., Farfade et al., 2015; Sun et al., 2014; Taigman et al., 2014). This success was facilitated by advances in deep learning and the compilation of large datasets of naturalistic face images (e.g., Cao et al., 2021; Huang et al., 2008; Kemelmacher-Shlizerman et al., 2016). For the particularly challenging task of face identification, it has been shown that face-id models perform equally well as forensic facial examiners (Phillips et al., 2018). A recent study by Dobs et al. (2022b) indicated that face-id models, rather than face-de or object-cat models, achieve human-level performance in an identity matching task. Notably, face-id models exhibited the most comparable representation of face stimuli to that of human face identification, as measured in behavioral multi-arrangement and similarity-matching tasks. To investigate whether humans and DCNNs achieve this similar performance by similar means, Abudarham et al. (2021) examined the role of critical features (e.g., eye shape, eyebrow thickness, and lip thickness) that have previously been isolated as critical features for humans as they can alter the perceived facial identity when manipulated (Abudarham & Yovel, 2016). Their results showed that face-id, but not object-cat models, use the same critical features as humans for face identification. In face-id models, the sensitivity to these features, as determined by the similarity between an original face example and versions with manipulated critical or noncritical features, correlated positively with behavioral face identification performance and computational identity selectivity towards late layers.

**Modeling face recognition with DCNNs**

Therefore, it can be noted that DCNNs display human-like performance in face recognition tasks, and there are reasons to believe that they process similar facial features to achieve this performance.

*Do humans and DCNNs display similar "psychological phenomena" of face recognition?*

Humans exhibit several well-documented "psychological phenomena" of face recognition, such as the "face inversion effect" (e.g., Yin, 1969), the "Thatcher effect" (e.g., Thompson, 1980), face familiarity effects (e.g., Johnston & Edmonds, 2009), and the "other-race effect" (e.g., Bothwell et al., 1989), which can be also tested in DCNNs. In this context, biological face recognition is often conceptualized as holistic template matching instead of purely atomistic feature processing (Freiwald et al., 2016; Tanaka & Farah, 1993). The "face inversion effect" leads to disproportionally reduced recognition performance and neural activity for inverted faces compared with inverted nonface objects (Kanwisher et al., 1998; Rossion & Gauthier, 2002; Yovel & Kanwisher, 2005). This effect was already reported in simpler computational models (Farzmahdi et al., 2016; Hosoya & Hyvärinen, 2017) and more recently in various DCNNs (Dobs et al., 2022b; Tian et al., 2022; Vinken et al., 2022; Xu et al., 2021; Yovel et al., 2022a; Zeman et al., 2022). While recent evidence suggests that the "face inversion effect" manifests only in face-id models (Dobs et al., 2022b), it has also been observed to a limited extent in face-de models (Tian et al., 2022; Xu et al., 2021), and in object-cat models that contained occasional faces (Vinken et al., 2022). The finding that DCNNs trained to simultaneously identify faces (i.e., different people) and nonface objects (i.e., different cups) misclassify and represent inverted faces as objects (Tian et al., 2022) could be also consistent with the hypothesis that upright and inverted faces are primarily processed by specialized face and object recognition systems in the brain (Haxby et al., 1999; Yovel & Kanwisher, 2005). Interestingly, inversion effects of nonface objects were reported to reach the level of faces, when the models are optimized for the corresponding identification task (Dobs et al., 2022b; Yovel et al., 2022a). These findings fit well with previously reported nonface inversion effects in human experts for categories such as dogs and birds (Campbell & Tanaka, 2018; Diamond & Carey, 1986). This implies that such inversion effects may be primarily driven by domain-specific mechanisms originating from the optimization for fine-grained expert tasks, which challenges the notion of the unique nature of facial stimuli.

**Modeling face recognition with DCNNs**

Strikingly, in DCNNs, these perceptual characteristics were even observed at the level of local facial features with the "Thatcher effect", in which inverted eyes and mouth are perceived as obvious distortions in upright but blend in well in inverted presentations. In this context, face-id but not object-cat models showed an increased representational dissimilarity between normal and Thatcherized versions, which was higher for upright than for inverted presentations and generally increased towards late layers (Jacob et al., 2021).

Thus, it can be argued that DCNNs, and face-id models in particular, exhibit indicators of established "psychological phenomena" of human face recognition.

*Does familiarity affect face identification in humans and DCNN similarly?*

Familiarity effects demonstrate that face identification is generally easier for humans in the context of familiar compared to unfamiliar identities (Young & Burton, 2018). As humans were reported to use the same critical features mentioned earlier for the identification of familiar and unfamiliar faces (Abudarham et al., 2019), this prompts merely conceptual rather than perceptual differences. In an approach to simulate familiarity in DCNNs, Blauch et al. (2021) tested face-id models before and after fine-tuning on new face identities. In face identification tasks on natural images, it was observed that familiarized face-id models surpass a simpler model, which combines principal component analysis (PCA) and linear discriminant analysis (LDA) to capture bottom-up and top-down information (Kramer et al., 2018). Young and Burton (2021) argued that the relatively simple PCA+LDA model already provided valuable insights and was only exceeded by the superscale performance of DCNNs. However, returning to the desired balance between computational abstraction and biological detail in neuroscientific models (Doerig et al., 2022), it is crucial to gather evidence for these effects also in more complex models, as they allow for a wider range of phenomena, questions, and explanations that cannot be addressed by simpler models (see Spotlights 1 and 2). The results by Blauch et al. (2021) highlighted that face experience in DCNNs increases their performance on unfamiliar faces but also their ability to learn novel identities. This supports the idea that face examples possess identity-general variability, which can be learned from other face configurations, but also identity-specific variability, which can only be learned through familiarization with the respective identity (see Spotlight 2). Consequently, face experience may enable rapid learning of identity-specific information and thereby enhance face identification for familiar faces. Another generalized form of familiarity effects in human face identification is the "other-race effect", which is characterized by a higher face identification performance for faces of familiar compared with unfamiliar ethnic groups. This effect was found in face-id models, based on the ethnicity of faces in the training sample (Tian et al., 2021), but not in face-de or object-cat models (Dobs et al., 2022b).

Therefore, it seems plausible that familiarity influences the ability of DCNNs to identify faces and may do so in a similar way as it does in humans. This broad spectrum of evidence suggests that DCNNs, and face-id models in particular, exhibit behavior similar to human face identification and, contrary to previous criticism (Bowers et al., 2022), may even account for various "psychological phenomena", such as the "face inversion effect", the "Thatcher effect", face familiarity effects, and the "other-race effect", under appropriate conditions.

## Computational level

At the computational level, comparisons between humans and DCNNs usually entail the investigation of encoded representations at the level of biological and artificial neurons, brain areas and model layers, and their processing mechanisms along the visual hierarchy.

*Do brains and DCNNs encode and process information for face recognition similarly?*

For the task of face detection, Ratan Murty et al. (2021) demonstrated that face-de models outperform descriptive models and human experts in predicting fMRI activity of face-, body-, and place-selective visual brain areas to novel images. Moreover, meaningfully trained models, deeper architectures with more layers or recurrent connections, and broadly sampled training data with different object categories all increased the fit to neural data, contrary to untrained models, shallower architectures with fewer layers, and narrowly sampled training data with different face identities only. Adequate DCNN predictions were observed for single voxels, generalized well across participants, and were confirmed in an additional high-throughput screening approach. The finding that face-selective artificial neurons are

**Modeling face recognition with DCNNs**

able to predict their biological counterparts supports the hypothesis that they may also use similar underlying features for face recognition. Nonetheless, the increased predictability of object-cat models trained on more diverse datasets reveals that, in the context of face detection, these mechanisms appear to depend to some extent on domain-general visual properties, as observed in related studies (Bao et al., 2020; Vinken et al., 2022).

For the task of face identification, deeper insights into the functional similarity of biological and artificial computations were given by Wang et al. (2022), who revealed that just like the brain (Cao et al., 2021; De Falco et al., 2016), face-id models develop single- and multiple-identity-selective neurons that generalize well to highly abstract examples. Similar to the findings on face selectivity (Ratan Murty et al., 2021), identity-selective artificial and biological neurons were found to encode information in similar ways through region-based feature coding (Chang & Tsao, 2017), that is, earlier regions/layers encode the axes of a face space and later regions/layers are specialized for specific regions of this face space and thus identity-selective. Subsequent analyses with selective lesioning of identity-selective artificial neurons revealed that artificial neurons tuned to single identities are more important for face identification than those tuned to multiple identities, which could also apply to their biological counterparts. Of course, unlike face-id models, the brain cannot be optimized for face identification alone, but must naturally solve other fine-grained expert tasks as well. This way, it could use the same domain-general or different domain-specific mechanisms for different expert tasks such as face and car identification. To test this with DCNNs, Kanwisher et al. (2023a) trained a fully-shared dual-task DCNN optimized for simultaneous face identification and object categorization, which has been shown to spontaneously develop functionally specialized face- and object-selective filters (Dobs et al., 2022a). Lesioning the top 20% of object-selective filters led to a more substantial deterioration in car identification performance than lesioning the same percentage of face-selective filters. Moreover, only a small portion of the most important car-selective filters overlapped with face-selective filters, while a more substantial portion of them overlapped with object-selective filters. The majority of car-selective artificial neurons were recycled from other previously nonselective artificial neurons. This provides computational evidence that fine-grained expertise is predominantly based on domain-specific

**Modeling face recognition with DCNNs**

mechanisms or task-specific optimization, rather than domain-general mechanisms. In addition, this implies that largely independently operating specialized processing systems could arise automatically in visual hierarchies that are optimized for different expert tasks (see Spotlight 1).

Accordingly, it can be argued that brains and DCNNs share similarities in the way they encode and process information for face recognition. Furthermore, these results provide preliminary evidence that face recognition may require predominantly domain-general mechanisms, whereas face identification may require predominantly domain- or task-specific mechanisms.

*Do brains and DCNNs encode features in a similar hierarchical fashion?*

The previously reported critical features (Abudarham et al., 2021; Abudarham et al., 2019) may incorporate several dimensions of face stimuli. In accordance with this idea, face-id models were reported to generalize identity representations well to novel encounters of altered intrinsic, face-related features (e.g., emotional expression) but not so much for those of changed extrinsic, face-unrelated features (e.g., head position and illumination; Xu et al., 2018). In addition, other studies indicated that face-id models predict human face identification best if they use similar 3-D shape features of face scans (Daube et al., 2021). A considerable number of studies underline the following hierarchical coding of facial features in DCNNs, which roughly corresponds to that observed in human and nonhuman primate brains (e.g., Freiwald, 2020; Freiwald & Tsao, 2010): **1.)** Similar to object recognition, early layers encode low-level features such as edges, blobs, orientation, and color (LeCun et al., 2015) that may be closely aligned with domain-general features. **2.)** Middle layers encode view-specific features such as head orientation and illumination (Grossman et al., 2019; Raman & Hosoya, 2020). **3.)** Late layers encode patterns of view-invariant features that are related to gender and identity (Abudarham et al., 2021; Jozwik et al., 2022; Parde et al., 2021; Raman & Hosoya, 2020; Wang et al., 2022). **4.)** The final layer encodes a nonlinear identity benefit for familiar faces (Blauch et al., 2021), similar to that observed in nonhuman primates in later processing areas outside the core face network (Landi & Freiwald, 2017; Landi et al., 2021).

However, there are also partly contradictory results. Raman and Hosoya (2020) compared face-selective areas in nonhuman primates and face-id model layers

**Modeling face recognition with DCNNs**

regarding various tuning properties (i.e., view-invariance, shape-appearance, facial geometry, and contrast polarity). In middle areas, no single layer was able to predict all combined tuning properties and only view-specific tuning was similar. Surprisingly, face-id, object-cat, and untrained models all indicated increasing view-invariant identity selectivity towards late layers. In contrast, Grossman et al. (2019) reported no significant correlations for late but only for middle layers. Here, artificial neurons encoded especially view-specific features. View-invariance and identity selectivity emerged towards late layers in both face-id and object-cat but not untrained models and were not organized in a brain-like face space. However, given a careful interpretation, these two specific studies could also represent two sides of the same coin by highlighting the view-specific, pictorial tuning in middle (Grossman et al., 2019) and view-invariant identity-tuning in late regions and layers (Raman & Hosoya, 2020), although their disagreement could be explained by the comparison of different species (i.e., humans and macaques) and/or models (i.e., VGGFace and AlexNet). Other studies also reported significant but very small correlations between face-selective brain areas and face-id models based on their representational similarity (Tsantani et al., 2021).

On the basis of the current literature, it can be stated that DCNNs should be recognized as sophisticated computational models of biological face recognition that, depending on the task, encode and process information in a hierarchical manner that largely resembles the neural face recognition system of humans and nonhuman primates. However, as biological face recognition is far more complex than simple object labeling and involves objectives beyond physical properties, such as those on social content (Oosterhof & Todorov, 2008; Sutherland et al., 2013), it is likely that image-computable models still fail to capture several functional properties (e.g., Jiahui et al., 2022; Tsantani et al., 2021) and yet represent the best current approximation of this ability.

## Spotlight 1: Face selectivity emerges through feedforward cascades

### Emerging selectivity

Debates on the origin of neural face selectivity hold that specialized face recognition may arise from nature (i.e., innate mechanisms), nurture (i.e., acquired experience), or the interaction of both. Studies with congenitally blind individuals indicated that face selectivity in occipitotemporal cortex may arise even in complete absence of visual experience (Ratan Murty et al., 2020; van den Hurk et al., 2017). Congenitally blind individuals were found to exhibit robust, functionally specialized neural responses to sounds associated with faces, bodies, places, and objects, which could be used to discriminate neural responses to the corresponding visual stimuli in sighted individuals (van den Hurk et al., 2017). Moreover, in this group of fully visually deprived individuals, such face-selective neural responses were observed even during haptic exploration of 3-D face stimuli (Ratan Murty et al., 2020). The striking similarities of functional specialization in congenitally blind and sighted individuals suggest that a category-selective map, including face selectivity and its topographic separation, may not require any visual inputs after all.

In this context, Xu et al. (2021) took advantage of the fact that architecture and training data can be separately controlled in DCNNs and tested whether domain-specific face experience is necessary for face selectivity to develop. Therefore, they created a fully face-deprived object-cat model by carefully removing all images that included occasional face stimuli (e.g., in the image background) from the model's training dataset. Interestingly, the face-deprived model performed only slightly worse in face detection and identification compared with standard face-de and face-id models. The results indicated that face-selective artificial neurons automatically emerged but were reduced in overall selectivity and sparseness. This automatic emergence of face selectivity in face-deprived models seems to rule out the need for acquired domain-specific face experience and supports previous evidence in nonhuman primates showing that face selectivity can develop automatically without domain-specific face experience (Sugita, 2008). Consistent with these findings, Vinken et al. (2022) showed that, in face patches of nonhuman primates, neural tuning

**Modeling face recognition with DCNNs**

for nonface objects is more predictive of face presence than neural tuning for faces itself. This reinforces the aforementioned assumption that face selectivity is predominantly based on domain-general features that are highly correlated with the category of faces. While the results made it clear that an object-cat model is also able to predict neuronal face selectivity in face patches, the highest similarity was surprisingly found in middle layers. This observation indicates that the reliable predictability of nonface object tuning for face presence could depend predominantly on local part-based features encoded in face patches. This aligns with the concept that face selectivity underlying face detection may arise either from the statistical properties of feedforward hierarchies, as in the case of purely feedforward DCNNs, or from acquired domain-general object experience.

To further disentangle these two possibilities, Baek et al. (2021) investigated untrained models and reported similar emergence of face- and object-selective artificial neurons that shared many characteristics with biological neurons of nonhuman primates. In line with the aforementioned findings by Xu et al. (2021), the automatically emerging face selectivity increased across layers and was higher for global compared to local face configurations, whereas face experience sharpened face-selective artificial neurons and thereby increased their sparseness. Similar to previous work indicating that untrained models are able to perform object recognition (Jarrett et al., 2009; Pinto et al., 2009), this automatically emerging face selectivity could serve as the basis for successful face detection. More precisely, this implies that elementary face selectivity could potentially manifest even without visual experience through the brain's feedforward circuitry composed of simple and complex cells. Although the exact computational mechanisms for this phenomenon are not yet well understood and require further exploration, initial studies already point to several promising but ambiguous explanations. In any case, this automatically arising face selectivity could serve as a catalyst for later fine-tuning through domain-general and domain-specific experience to enable more demanding tasks such as face identification.

In accordance with this reasoning, object-cat models were reported to provide a better foundation for subsequent learning of a fine-grained identification of cars (Kanwisher et al., 2023a) and birds (Yovel et al., 2022b) than face-id models. The inability of face-id models to adapt well to new expert tasks of different domains was

**Modeling face recognition with DCNNs**

found for both superordinate (i.e., related to bird species) and individual (i.e., related to bird identity) levels (Yovel et al., 2022b). This corroborates the idea that fine-grained expertise, as in the case of face identification, may rely primarily on domain-specific mechanisms or task-specific optimization but not just domain-general mechanisms or the "specialness" of face stimuli.

## Disentangling selectivity

To investigate how functional specialization with multiple selective systems could arise, Dobs et al. (2022a) trained a face-id and object-cat model each in a single-task condition and a fully-shared dual-task model with interleaved batches of faces and objects. As expected, single-task models performed worse on the respective other task, whereas the dual-task version succeeded in both tasks. Interestingly, however, a functional specialization with face- and object-selective artificial neurons arose automatically in dual-task models after middle layers and increased steadily towards late layers. Through lesioning these selective artificial neurons, this unsupervised task segregation was then confirmed a double dissociation that emerged only for meaningfully separated tasks and to a smaller degree for other nonface categories. This further suggests that some tasks (e.g., faces vs. objects) segregate better than others (e.g., food vs. objects), and hence the former may rely on more distinct feature sets compared with the latter.

Although feedforward hierarchies in DCNNs enable the emergence of category-selective artificial neurons, this selectivity is not organized in spatial clusters as in functionally specialized brain areas. This is because invariant object recognition in these models is tied to the selectivity of entire convolutional feature maps rather than specific retinotopic receptive fields. However, techniques such as self-organizing maps (SOMs; Kohonen, 1990), that is, unsupervised artificial neural networks that learn to represent high-dimensional data in a low-dimensional space, or additional loss functions, for example, simulating the wiring costs between the artificial neurons of DCNNs, can be used to visualize and investigate the resulting topographies. In such an approach, Cowell and Cottrell (2013) used SOMs to replicate object-selective topographies reported in occipitotemporal cortex using fMRI (Haxby et al., 2001;

**Modeling face recognition with DCNNs**

Spiridon & Kanwisher, 2002). Similar to recent evidence highlighting the importance of domain-general mechanisms in the development of face selectivity (Bao et al., 2020; Vinken et al., 2022), their results suggested that a functionally specialized topography may also arise automatically through such mechanisms. Along these lines, the particularly high selectivity for faces could be because of their high within-category and low between-category similarity. These findings are further supported by recent unsupervised approaches, such as DCNN representations fed into SOMs (Doshi & Konkle, 2023) and topographic variational autoencoders (Keller et al., 2021), which produce similar topographic maps. Although domain-general mechanisms may be sufficient to generate this object-selective topography in SOMs, this observation appears to be limited to face detection only, and arguably oversimplifies this phenomenon due to the models' assumption of a limited set of visual parameters and lack of consideration of hierarchical interactions (Blauch et al., 2022).

To expand on this idea, Blauch et al. (2022) examined interactive topographic networks that incorporate biologically plausible constraints. These networks were composed of object-cat models with additional recurrent layers, optimized to balance task performance and wiring costs in order to account for dependencies within the topographic map. As expected, these models displayed face-, object-, and scene-selective patches with columnar responses and efficient connectivity. Intriguingly, among various tested mechanisms, recurrence, separated excitatory and inhibitory artificial neurons, as well as purely excitatory feedforward connections, produced the most pronounced selective clusters. In another study, Lee et al. (2020) implemented a similar topographic DCNN that was optimized for object categorization while simultaneously minimizing the wiring cost of its artificial neurons. The resulting object-selective topography strongly resembled the functionally specialized organization of primate inferior temporal cortex, as especially face- and body-selective artificial neurons were organized in adjacent or overlapping clusters, and especially face-selective clusters were strongly interconnected to form a network across model layers.

To summarize, recent studies investigating face selectivity in DCNNs strengthen a computational account for the emergence of elementary face selectivity through neither purely innate mechanisms (i.e., hardwired from the beginning) nor specific experience (i.e., acquired later on) but rather the statistical properties of feedforward hierarchies. The development of face-selective neurons could be guided by additional selection pressure to categorize objects and minimize wiring effort, leading to their characteristic topographic organization.

# Spotlight 2: Face identification is enhanced by identity-specific experience and generative mechanisms

## Identity-specific experience

An important property of biological face recognition is its great robustness in challenging conditions. Especially familiarity is thought to enable humans to identify faces even under the most difficult conditions. In this regard, Noyes et al. (2021) demonstrated that the performance of face-id models decreases similarly to that of humans in the presence of deliberate disguise (Noyes & Jenkins, 2019). In face-id models, "face averaging" (i.e., averaging the representations of multiple examples of a given identity), which was proposed as a computational mechanism for biological face identification (Burton et al., 2005; Kramer et al., 2015), led to enhanced within-identity matching in evasion disguise (i.e., correctly identifying the same identity even when the person is disguised to look nothing like themselves), but reduced between-identity matching in impersonation disguise (i.e., correctly distinguishing different identities, even though one is disguised to look like the other). Curiously, performance on both within-identity and between-identity matching was enhanced by a contrast learning approach that emphasized differences in appearance and identity between individuals. This suggests that averaging multiple perceived face examples of a given identity may lead to a more robust representation for subsequent identification under changing or more difficult conditions.

## DCNNs and generative models

As DCNNs are gradually entering face recognition research, the question arises whether they are indeed more suitable than preceding statistical or generative models (hereafter GMs; Tolba et al., 2006). As an example, Jozwik et al. (2022) reported that face-id models do not predict human identity judgements significantly better than the

**Modeling face recognition with DCNNs**

*Basel Face Model* (BFM; Gerig et al., 2018; Paysan et al., 2009), a well-known 3-D morphable model based on principal components of 3-D face scans. Interestingly, both DCNNs trained on either natural images or generative examples predicted human judgements equally well. Although this may not surprise as both DCNNs and the BFM were tested on face stimuli that were generated by the BFM, this could also emphasize the ecological validity of this statistical tuning in both in vivo and in silico face spaces. These findings are also encouraged by comparisons with a similar 2-D morphable model, the widely used *Active Appearance Model* (AAM; Cootes et al., 2001; Edwards et al., 1998). The AAM was found to explain face-selective biological neurons equally well and even better in terms of identity-unrelated, extrinsic features (e.g., illumination) compared with face-id models (Chang et al., 2021).

It is worth noting that the previously mentioned studies also used artificially generated face stimuli for testing, eliminating a key advantage of DCNNs. As reported by O'Toole et al. (2018), DCNNs and GMs display differently constructed face spaces. Similar to how biological face recognition can result in ambiguous face representations because of factors such as familiarity or viewing conditions, DCNNs encode not only the identity but also the quality of images in their face space. Thereby, low-quality and hard-to-recognize examples are positioned in the center of their face space, whereas higher-quality and easier-to-recognize examples are positioned at the outer edges. In contrast, GMs represent the prototypical face in the center of their face space because image quality is usually controlled here. This ability of DCNNs to learn and represent a variety of complex features directly from the data is invaluable in studies aimed primarily at examining face recognition processes in naturalistic settings. Because GMs can learn the underlying data distribution and generate new samples, they are particularly appealing in scenarios where there are only a few training examples. Accordingly, the combination of DCNNs and GMs could be particularly promising.

**Analysis-by-synthesis**

Novel insights regarding the underlying mechanisms of face identification were recently obtained through a combination of DCNNs and GMs. Yildirim et al. (2020) implemented an efficient inverse graphics model, which combines a face-id model and

**Modeling face recognition with DCNNs**

the BFM. Instead of mapping a face image directly onto an identity label, this efficient inverse graphics model uses DCNN and GM components to go the other way around. First, a DCNN component is used to perform preliminary segmentation and normalization of faces, as well as to extract latent 3-D variables (e.g., shape, texture, and viewing parameters), and to perform standard face identification. Then, a GM component generates a 3-D representation based on the extracted latent variables. Finally, this representation is used to create a 2.5-D surface representation, followed by a 2-D segmented face image. This way, in contrast to common face-id models, this analysis-by-synthesis network not only analyzes a 2-D image and outputs the corresponding identity label (e.g., "Albert Einstein"), but instead feeds a generative model with facial parameters to infer a 3-D identity representation (e.g., a 3-D model of Albert Einstein's face), which is then used for identification. The model thereby infers parts of the face from the image, which cannot be directly extracted. The efficient inverse graphics model demonstrated remarkable performance by achieving an exceptional level of similarity to neural data of nonhuman primates ($r = .96$) at corresponding processing stages of the model and face patches with various hallmark features of face coding (such as view specificity, mirror symmetry, and identity specificity coding), when compared with standard DCNNs ($r = .36$), and thereby seemed to close some of the previously reported gaps in capturing the full hierarchy of biological face recognition (Grossman et al., 2019).

To summarize, recent studies investigating face identification in DCNNs indicate that this ability is enabled through domain-specific face experience for unfamiliar identities, that is, clustering inter-identity examples in a face space. Face identification is further enhanced through identity-specific experience for familiar faces, that is, clustering intra-identity examples closer together. Moreover, additional generative mechanisms may enable inferences that boost this capacity even more. In biological face recognition, these generative mechanisms could be found in recurrent and feedback connections.

# Conclusion

In this review, we have summarized recent studies on modeling biological face recognition with DCNNs. On the basis of these findings, we conclude that they are useful models for studying biological face recognition under the right circumstances. To begin with, DCNNs are largely able to exhibit face recognition and identification behaviors similar to humans, and even account for several "psychological phenomena" from face recognition research such as the "face inversion effect", the "Thatcher effect", face familiarity effects, and the "other-race effect". In addition, DCNNs seem to encode and process face information in a highly similar manner compared to the ventral visual pathway and core face network of humans and nonhuman primates. This synthesis is consistent with another recent review on this topic by O'Toole and Castillo (2021), which describes four technical stages of face recognition learning in DCNNs: The first stage involves between-category learning for object categorization, which defines domain-general object experience. The second and third stages involve within-category learning for face identification and adaptation to the specific characteristics of individuals and environments, which describe domain-specific face experience. The fourth stage of learning individual people corresponds to identity-specific experience. Crucially, the two spotlights presented add an additional stage before the first and after the final stage of this theoretical model (see Figure 2). This way, the foundation of face recognition learning in DCNNs may lie in the ability of feedforward hierarchies to enable the development of face selectivity even in the absence of visual experience, whereas the peak may consist of additional generative mechanisms that improve the effectiveness and robustness of face identification. Modeling biological face recognition with DCNNs has so far not only provided exciting novel insights, but its opportunity to control nature and nurture has opened up a new perspective on many levels for further investigation.

**Modeling face recognition with DCNNs**

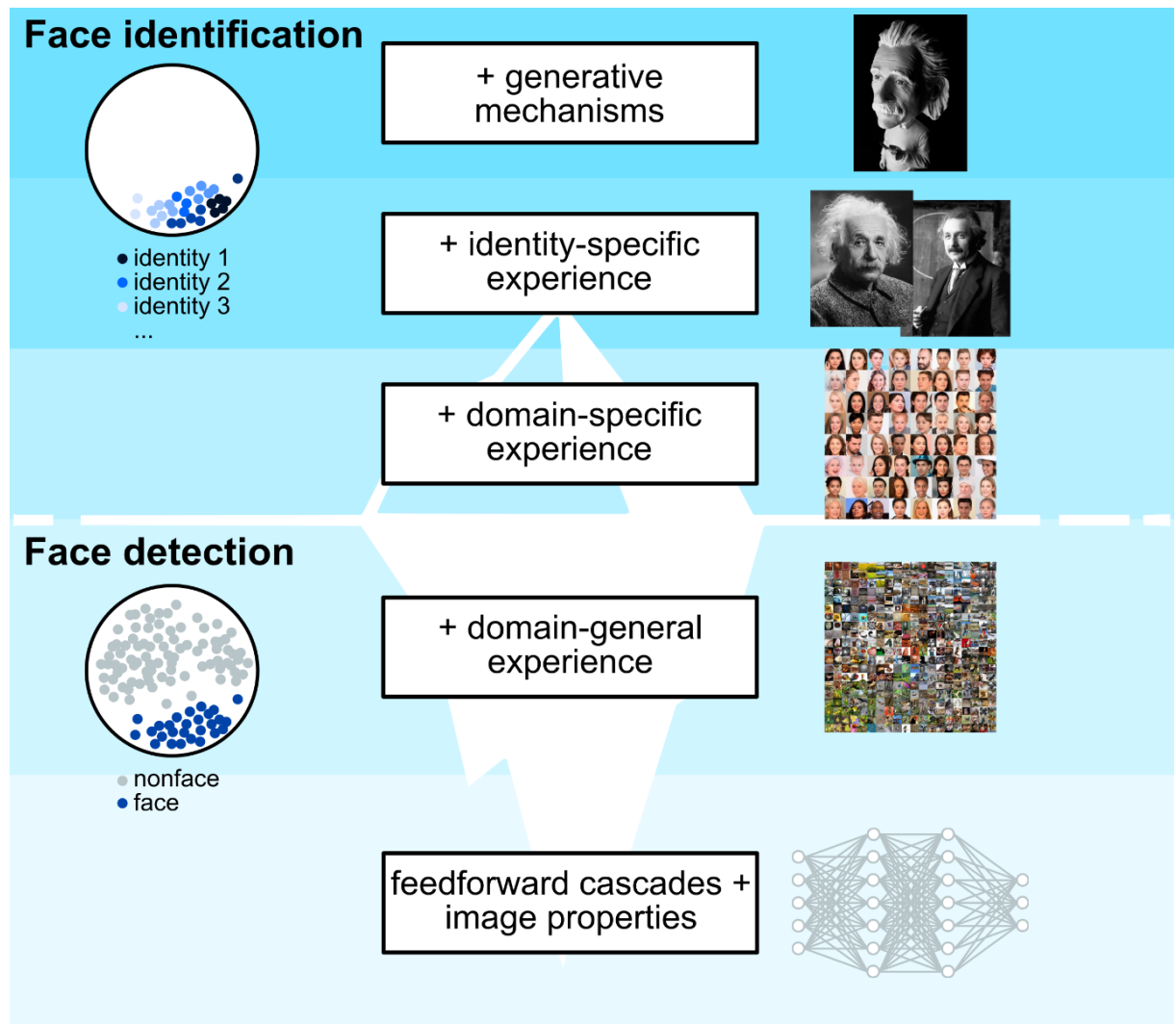

**Figure 2. Summary model of face recognition learning in DCNNs.** Elementary face selectivity, and hence face detection, arises automatically through feedforward processes without visual experience. Domain-general visual experience (i.e., various object categories) increases the ability to detect faces. Domain-specific visual experience (i.e., various face identities) further increases the ability to detect faces and serves as a basis for the ability to identify faces. Identity-specific experience (i.e., multiple examples of a given identity) enables robust face identification. Finally, this ability seems to peak with additional generative mechanisms, as the model is capable of inferring a 3-D representation from a 2-D image. **Source:** Images of objects obtained from the ImageNet database (Deng et al., 2009). To protect identities, the displayed non-famous faces were generated by an artificial neural network (https://thispersondoesnotexist.com/).

# Open questions and future directions

The previous successes from modeling biological object recognition with DCNNs should be transferred and extended to the field of biological face recognition. In particular, more causal approaches, such as the recent experiments regarding in silico lesioning analyses (e.g., Blauch et al., 2022; Dobs et al., 2022a; Kanwisher et al., 2023a; Wang et al., 2022), deprivation studies (e.g., Baek et al., 2021; Xu et al., 2021), and the synthesis of maximal activations (e.g., Ratan Murty et al., 2021; Vinken et al., 2022), should be further expanded to refute the criticisms of "prediction-based experiments" and to design primarily "controlled experiments" with adequate alternative explanations (Bowers et al., 2022). To shape this development in a targeted manner, it seems particularly important that future studies report aspects such as applied architectures, datasets, tasks, and hyperparameters with greater detail and transparency (see Table 1). On the one side, the advantage that architecture can be fully controlled in DCNNs should be further exploited. Instead of performing the same analyses with a similar model to control for differences in architecture, it might be helpful to also include particularly dissimilar models that should not demonstrate a given phenomenon. The addition of, for example, generative mechanisms, recurrent connections, biologically plausible connectivity, attentional mechanisms, and unsupervised learning could yield new insights about the influences of architecture. On the other side, control over the dataset should also be further leveraged. Although more comprehensive and naturalistic datasets have the potential for better results (e.g., Allen et al., 2022; Hebart et al., 2019), the addition of more creative datasets, such as those used in recent work on fine-grained identification of nonface categories (Dobs et al., 2022b; Yovel et al., 2022b), and customized datasets, such as artificial "controversial stimuli" created to reveal model preferences in synthesized images of mixed categories (Golan et al., 2020), could further strengthen the results and should be included.

Another promising line of future research involves comparisons with other measurements and populations. Although to date most findings are based on comparisons with fMRI data, other measurement techniques should be incorporated. As an example, comparisons with electrophysiological data could help to better understand how the representations encoded in the brain at different time points map

**Modeling face recognition with DCNNs**

to different model layers (e.g., Cichy et al., 2016) in the case of face recognition. As another example, comparisons of eye tracking data and model saliency maps could also shed more light on the attentional mechanisms (e.g., van Dyck et al., 2022) underlying face detection and identification. Furthermore, as developmental studies showed that the gaze behavior of 4-month-old infants is more consistent with earlier model layers, whereas that of 12-month-old infants is more consistent with later layers (Kiat et al., 2022), this suggested gradual shift towards greater abstraction may be especially interesting in the case of face recognition. Similarly, recent studies in young infants have shown that face selectivity develops early on (Kosakowski et al., 2022). Therefore, longitudinal studies could be used to link the stages of face recognition learning in the infant brain with those observed in DCNNs. In addition, comparisons involving patients with impaired face recognition or other species (e.g., monkeys or rodents) might be useful to extend the robustness of the current findings.

As previous studies have focused primarily on the physical properties of biological face recognition, future studies may also attempt to illuminate the functional properties. Initial studies with this aim have already demonstrated that social aspects of face recognition such as facial expression (Colón et al., 2021) and social judgements (Keles et al., 2021) may already be partially encoded by associative learning in current face-id models.

Finally, it is worth noting that face recognition in real-world applications, such as those for security, biometric authentication, marketing, and healthcare, is much more challenging than in closely controlled experiments. Although the studies summarized in this review offer exciting insights for basic research at the intersection of neuroscience and artificial intelligence, the difficulty often lies in their practical implementation (for more technical perspectives, see Guo et al., 2020; Li et al., 2015).

# Modeling face recognition with DCNNs

**Table 1. List of reported studies that compared biological face recognition and DCNNs.**

| Study | DCNN model family | Training dataset | Training task | Testing dataset | Testing task |
|---|---|---|---|---|---|
| Abudarham et al. (2021) | GoogLeNet, VGG | ImageNet, VGGFace2 | object-cat, face-id | custom | face-id |
| Abudarham et al. (2019) | OpenFace | CASIA-WebFace, FaceScrub | face-id | custom | face-id |
| Baek et al. (2021) | AlexNet | none | untrained | ImageNet, VGGFace2 | encoding |
| Blauch et al. (2022) | interactive topographic network (based on ResNet) | ImageNet, VGGFace2, Places365 | object-cat + face-id + scene-cat | - | face-id |
| Blauch et al. (2021) | VGG | ImageNet, LFIW, VGGFace2 | untrained, object-cat, face-id | LFIW, VGGFace2 | face-id |
| Chang et al. (2021) | AlexNet, CORnet, VGG | ImageNet, VGGFace2 | object-cat, face-id | various | face-id |
| Colón et al. (2021) | ResNet | Universe | face-id | Karolinska Directed Emotional Faces | encoding |
| Daube et al. (2021) | ResNet (various objective functions) | artificially generated face stimuli | various | artificially generated face stimuli | face-id |
| Dobs et al. (2022a) | VGG | ImageNet, VGGFace2 | object-cat, face-id, dual-task | various | object-cat, face-id |
| Dobs et al. (2022b)[a] | AlexNet, VGG | ImageNet, VGGFace2, others | untrained, object-cat, face-de, face-id | LFIW, VGGFace2 | encoding |
| Doshi and Konkle (2023)[a] | AlexNet, SOMs | ImageNet | object-cat | various | encoding |

**Modeling face recognition with DCNNs**

| | | | | | |
|---|---|---|---|---|---|
| Grossman et al. (2019) | VGG | ImageNet, VGGFace2 | object-cat, face-id | various | encoding |
| Jacob et al. (2021) | VGG | ImageNet, VGGFace2 | object-cat, face-id | IISc Indian Face Dataset | encoding, "Thatcher effect" |
| Jiahui et al. (2022)[a] | AlexNet, ResNet, VGG | ImageNet, MS-Celeb-1M | object-cat, face-id | LFIW, custom | face-id, encoding |
| Jozwik et al. (2022) | AlexNet, VGG | artificially generated faces | various tasks | artificially generated faces | encoding, similarity judgements |
| Kanwisher et al. (2023a) | VGG | ImageNet, VGGFace2 | object-cat, face-id, car identification[b] | CompCars | car identification |
| Keles et al. (2021) | ResNet, VGG | Chicago Face Database, ImageNet, others | face-id | various | encoding, social judgements |
| Lee et al. (2020)[a] | topographic DCNNs (based on AlexNet) | ImageNet, LFIW, Places365, Open Images | object-cat | various | encoding |
| Noyes et al. (2021) | Sankaranarayanan et al. (2016) | CASIA-WebFace | face-id | FAÇADE | face-id |
| Parde et al. (2021) | ResNet | Universe | face-id | IARPA Janus Benchmark-C | face-id |
| Phillips et al. (2018) | various | various | face-id | custom | face-id |
| Raman and Hosoya (2020) | AlexNet, VGG | VGGFace2, (ImageNet, Oxford-102) | face-id | FEI face database, custom | encoding |
| Ratan Murty et al. (2021) | various | ImageNet, Places2, VGGFace2 | object-cat, scene-cat, face-id | THINGS | face-de |
| Tian et al. (2022) | AlexNet, VGG | ImageNet, VGG-Face | object-cat, face-id | CASIA-WebFace, custom | encoding, "face inversion effect" |

**Modeling face recognition with DCNNs**

| Tian et al. (2021) | VGG | VGG-Face | face-id | VGGFace2 | face-id, "other-race effect" |
|---|---|---|---|---|---|
| Tsantani et al. (2021) | OpenFace | CASIA-WebFace, FaceScrub | face-id | custom | face-id |
| Vinken et al. (2022)[a] | Alexnet, GoogLeNet, ResNet, VGG | ImageNet | object-cat | custom | encoding |
| Wang et al. (2022) | VGG | CelebA, VGGFace2, (ImageNet) | face-id | custom | face-id |
| Xu et al. (2018) | ResNet | artificially generated face stimuli | face-id | artificially generated face stimuli | face-id |
| Xu et al. (2021) | AlexNet, ResNet | ImageNet, (Faces in the Wild, CASIA-WebFace) | object-cat | custom | face-de, face-id |
| Yildirim et al. (2020) | efficient inverse graphics model (based on AlexNet), VGG | artificially generated faces | face-id | face-identities-view | face-id |
| Yovel et al. (2022a)[a] | VGG | ImageNet, VGGFace2, others | object-cat, face-id, bird identification[b] | custom | encoding, "face inversion effect" |
| Yovel et al. (2022b)[a] | VGG | ImageNet, VGGFace2, others | object-cat, face-id bird identification[b] | custom | bird identification |
| Zeman et al. (2022)[a] | AlexNet, DeepFace | ImageNet, CASIA-WebFace | object-cat, face-id | Mooney face online dataset | face-de, "face inversion effect" |

*Note:* [a] = Preprint at the time of writing, [b] = Different fine-tuning.

**Modeling face recognition with DCNNs**

Modeling face recognition with DCNNs31

**Modeling face recognition with DCNNs**

**Modeling face recognition with DCNNs**

**Modeling face recognition with DCNNs**

**Modeling face recognition with DCNNs**

**Modeling face recognition with DCNNs**

**Modeling face recognition with DCNNs**

**Modeling face recognition with DCNNs**

**Modeling face recognition with DCNNs**

**Modeling face recognition with DCNNs**